# Highlights

## Multi-class Categorization of Reasons behind Mental Disturbance in Long Texts
Muskan Garg

- Causal Analysis: From inferring mental states to finding causal indicators in self-reported text.
- Introduce a Multi-task CAMS dataset for causal analysis.
- Deploying Longformer to handle long text for multi-class causal categorization.
- Perform comparative analysis of tranformer-based models and cause-based analysis for causal categorization.
- Interpret the effectiveness of rule-based discourse analysis using ablation studies.

# Multi-class Categorization of Reasons behind Mental Disturbance in Long Texts

Muskan Garg[a,*]

[a]*Mayo Clinic, Rochester, 55905 MN, United States of America*

ABSTRACT

Motivated with recent advances in inferring users' mental state in social media posts, we identify and formulate the problem of finding causal indicators behind mental illness in self-reported text. In the past, we witness the presence of rule-based studies for causal explanation analysis on curated Facebook data. The investigation on transformer-based model for multi-class causal categorization in Reddit posts point to a problem of using long-text which contains as many as 4000 words. Developing end-to-end transformer-based models subject to the limitation of maximum-length in a given instance. To handle this problem, we use Longformer and deploy its encoding on transformer-based classifier. The experimental results show that Longformer achieves new state-of-the-art results on M-CAMS, a publicly available dataset with 62% F1-score. Cause-specific analysis and ablation study prove the effectiveness of Longformer. We believe our work facilitates causal analysis of depression and suicide risk on social media data, and shows potential for application on other mental health conditions.

## 1. Introduction

As per reports released in August 2021[1], *1.6 million people* in England were on waiting lists for mental health care. As per estimation, *8 million people* could not get specialist help as they were not considered *sick enough* to qualify. This situation underscores the need for automation of mental health detection from social media data where people express themselves and their thoughts, beliefs/ emotions with ease. This self-reported social media data is valuable but laborious for manual interpretations; thus, although complex, an automated system would significantly enhance the ability to understand a social media user's state of mental health.

Amid COVID-19, the social NLP research community witness increase in the use of social media to express thoughts/ feelings and share life experiences Gianfredi, Provenzano and Santangelo (2021). Mental health has thus become one of the most challenging areas in social well-being. Social NLP research community uses social media platforms such as Twitter, Reddit, Facebook and Gab to find user profiles at-risk. We move one step ahead of classifying such vulnerable posts to find possible reasons behind users' mental disorders in their writings.

We witness a variety of reasons from life experiences to cause mental disturbance due to their perception towards life and circumstances. Being a deep philosophical topic of discussion, clinicians have mutual consensus on increased complexity of thoughts involved in examining mental disorders Roessner, Rothe, Kohls, Schomerus, Ehrlich and Beste (2021). Additionally, limited availability of clinical psychologists and therapists has made *mental health detection and analysis* even more challenging Coffman, Bates, Geyn and Spetz (2018). To this end, we propose the NLP-centered approach to find indicators behind mental health degradation in self-reported texts of social media users.

Consider a post *P* in a social media platform - Reddit posted in subreddit *r/depression*. The post *P* is personally written by a user which exhibit higher levels of emotions and stances associated with mental and social well-being, respectively. The post *P* written by user is given as:

> *P* = "I do not want to read literature but my parents forced me to do so. Not happy with my grades"

A major concern of user is about *education* stating the issue of forced subjects by parents that clearly indicate child's lack of interest affecting state of mind. The writer considers the effect of low grades hindering the opportunities in *jobs and career* which becomes a potential reason behind degradation of mental stability. Henceforth, the text segments such as *parents forced* and *not happy with grades* points to assign a category for causing mental disturbance as *jobs and career*. The complexity of problem underlies in the fact that this post *P* is related to *jobs/ career* (affecting grades in university) and *relationship* (parents enforcing subjects resulting in bitterness in relations). However, to resolve such perplexity, the perplexity guidelines suggest the *root cause* as prime reason Garg, Saxena, Krishnan, Joshi, Saha, Mago and Dorr (2022). In upcoming section, we define the problem of causal analysis for mental health domain.

### 1.1. Problem Definition

Being a general term, *causal analysis* refers to the field of experimental design and statistics pertaining to established cause-and-effect relationship.[2] We advocate the recent surge in *causal analysis* for NLP-centered applications and computer vision.[3] In this section, we define causal analysis for mental health domain as *cause-and-effect relationship* in user-generated texts for underlying cause-based categorization.

---

*Corresponding author
muskanphd@gmail.com (M. Garg)
www.drmuskan.in (M. Garg)
ORCID(s):

[1]https://www.theguardian.com/society/2021/aug/29/strain-on-mental-health-care-leaves-8m-people-without-help-say-nhs-leaders

[2]https://en.wikipedia.org/wiki/Causal_analysis

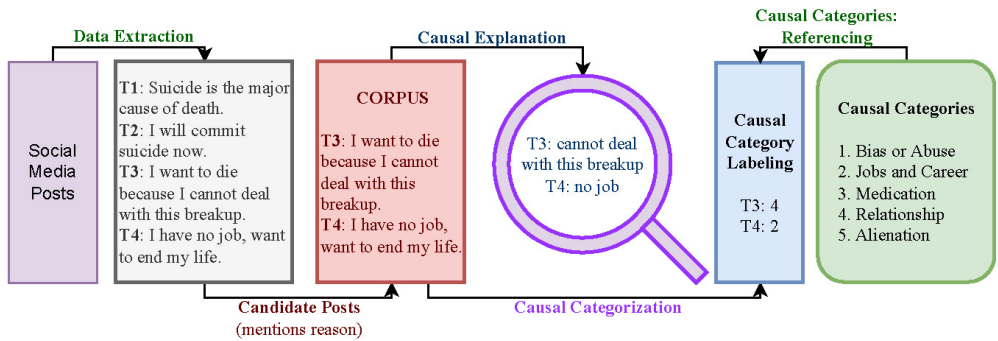

**Figure 1:** Overview of the problem statement - 'Causal Analysis of Mental disturbance in Social media posts'. In this work, we solve *causal category labeling* in a given *corpus* of candidate posts by referring pre-determines five causal categories.

Cognitive functions of a human brain enable perceiving, understanding and reflecting information over prevailing circumstances. The complexity of users' perception (belief, values and morals) may cause stressful situations. Long term stress, for weeks or months, may intensify it to clinical depression. Prevailing depression for months and years may lead to an extreme case of cognitive mental decline which affects social well-being. To avoid cognitive decline, cause detection in social media posts may therefore contribute towards mitigating mental wellness.

The effects of cognitive decline points to *clinical depression* and *suicidal tendencies*. The self-advocacy of users' experiences may contain indicators of cause behind these effects. NLP-based analysis make such inferences to find cause-and-effect relationship in a user-generated text. Being a highly complex problem of finding cause behind mental disorders, it is difficult to find reasons in a given text. We pin down five major categories to simplify complexity of the problem. In upcoming section, we formulate the problem to kick-start a project for building efficient models on causal analysis as a potential research direction.

### 1.2. Problem Statement

Causal analysis for mental illness is three-fold: (i) cause detection, (ii) causal explanation, and (iii) causal categorization. In *cause detection*, posts that does not indicate any reasons are filtered out to obtain candidate posts as evident from Figure 1. For instance, consider the following posts $T1$ and $T2$:

$T1$: Suicide is the major cause of death.
$T2$: I will commit suicide now.

As $T1$ mention a general fact and $T2$ is a direct statement about intent of a user, we find no-mention of any reason. Such posts are filtered out to retain other candidate posts that mention potential cause behind mental disturbance.

In this work, we use these candidate posts as corpus for causal categorization. The *causal explanation* extraction is a task of finding text segments leveraging reasons in a given data point. Such text segments facilitates labeling category which is defined as *causal categorization*. As mentioned before, we select five causal categories to classify a given instance in association with discussions carried out by a social NLP researcher, a clinical psychologist and a rehabilitation counselor. Interdisciplinary studies with clinical psychology theories and NLP-based analysis of social media data suggests five causal categories as (i) bias or abuse, (ii) jobs and career, (iii) medication[4], (iv) relationship, and (v) alienation.

To the best of our knowledge, there are no existing dataset for identifying reason behind mental illness in self-advocated texts. We build **M**ulti-tasking in **C**ausal **A**nalysis of **M**ental health illness in **S**ocial media posts (M-CAMS) dataset.[5] For each causal category, a sample is given in Table 1 in the format of <text, causal detection (CD), causal categorization (C. Category), causal explanation (C. Explanation)>. We formulate the problem of causal categorization as multi-class classification problem on user-generated long texts.

### 1.3. Challenges

Reddit, a social media platform, is divided into thousands of smaller communities called subreddits, such as *r/depression* and *r/SuicideWatch* to facilitate Reddit users posting their life experiences. For Reddit data, the limit on number of characters for a comment is 10,000, for a post is 40,000 and for post titles is 300 characters. Being end-to-end model, transformer considers a maximum token length of 300-400 words in a given text. They truncate long texts to maximally supported length of first 300-400 word-pieces. To this end, we determine the length of each Reddit post in a corpus and examine it as shown in Table 2. We find inconsistency in the length of a post in the range of [1, 4378]. We further examine the percentage of posts with more than 200, 300 and 400 words in respective categories and found about 32.57%, 18.27% and 11.96% of posts, respectively. The minimum, maximum and average length of candidate posts for each causal category indicates the need of better classifiers which are applicable to long texts. In this paper, we deploy transformer based models for multi-class classification of long texts.

---
[3]https://github.com/fulifeng/Causal_Reading_Group

[4]We recognize 'medication' as both an *indicator* of physical/emotional illness (e.g., an intent to alleviate illness) and a potential *cause* of illness (e.g., medication-induced depression).

[5]https://github.com/drmuskangarg/CAMS

| Text | CD | C. Category | C. Explanation |
|---|---|---|---|
| That's all I can really say. Nothing is worth the effort... I don't think I am capable of taking steps to improve my life, because I just don't even fucking care. Why try... I just... ugh... | 1 | Alienation | Nothing is worth the effort |
| God help me.... I know I should go to the hospital. I know I have to keep fighting....if only to prove to my children, cursed with these genetic tendencies of mine, that life is worth living. My scars and cynicism are just a little too hard for anyone who tries to stay around too long. | 1 | Medication | go to the hospital, scars and cynicism, genetic tendencies |
| I hate my job .. I cant stand living with my dad.. Im afraid to apply to any developer jobs or show my skills off to employers..I dont even have a car rn... I just feel like a failure. | 1 | Jobs and Careers | hate my job, feel like failure |
| 5 of my closest friends from high school have stopped responding to my calls or texts. i thought it was just a phone issue at first, but it is too unlikely of just coincidence. | 1 | Relationships | 5 of my closest friends, stopped responding |
| ...Then, on the way to the pub, a group of girls basically called me unattractive. Funny how girls are never shy about calling me ugly, but they're apparently too shy to "approach me". | 1 | Bias or Abuse | girls call me unattractive |
| Does anyone feel like the only person that could understand your depression would be someone else that was depressed? It might suck you into a place that you don't want to be in again. | 0 | - | - |

**Table 1**
Sample of M-CAMS dataset for causal analysis that follows the format of <text, causal detection (C. Detection), causal categorization (C. Category), causal explanation (C. Explanation)>

| Causal Category ↓ | Total #(Posts) | Min. L | Max. L | Avg. L | > 200 | > 300 | > 400 |
|---|---|---|---|---|---|---|---|
| Bias or Abuse | 350 | 5 | 4378 | 296.56 | 39.71 | 23.14 | 16.57 |
| Jobs and career | 628 | 13 | 2771 | 234.92 | 40.28 | 22.61 | 14.33 |
| Medication | 623 | 3 | 3127 | 207.83 | 30.65 | 19.74 | 13.96 |
| Relationship | 1342 | 2 | 3877 | 230.57 | 38.31 | 23.06 | 14.36 |
| Alienation | 1408 | 1 | 1592 | 151.19 | 22.72 | 11.29 | 6.61 |

**Table 2**
Word length variation in Reddit posts across causal classes in the corpus of candidate posts where each category contains instances with minimum length (Min. L), maximum length (Max. L) and average length (Avg. L). Length determines the number of words in a given text. The percentage of posts in a corresponding category having > 200 words, > 300 words and > 400 words suggest AI models for long text classification.

### 1.4. Contributions

A CAMS dataset contains six categories where the sixth one belongs to *no reason* implying no cause-and-effect relationship among text segments which suggests filtering out these irrelevant posts as a separate task of *cause detection*, before *causal categorization*. As a result, introduction of M-CAMS dataset have potential to capture multi-tasking behaviour for causal analysis. In this work, we focus on causal categorization with major contributions as:

1. Introducing M-CAMS dataset for the task of cause detection, causal explanation, and causal categorization in social media posts for mental disturbances.
2. Formulate and solve the problem of multi-class causal categorization of mental disturbances by simplifying the complexity of task in five causal categories.
3. Deploy transformer-based multi-class classifiers for causal categorization in user-generated posts.
4. To test and validate the effectiveness of transformers and rule-based discourse analysis through comparative analysis and ablation studies, respectively.

We further organise this paper in six different sections. Section 2 provides background through evolution of historical perspective and problem formulation. A detailed discussion on data collection, rule-based discourse analysis, and transformer based methods is given in Section 3. Furthermore, we brief out experimental setup, hyperparameter optimization, baselines, evaluation metrics in Section 4. We perform comparative analysis, cause-based analysis, computational analysis and ablation studies for in-depth performance evaluation of classifiers. Section 5 discusses ethical considerations and open challenges. Finally, Section 6 concludes the work.

### 2. Background

The social NLP research community has recently witnessed advancements in (i) *cross-section analysis*: classifying the type of mental disturbance from a given text, and (ii) *longitudinal analysis*: finding patterns in emotional spectrum present in historical timeline of social media users. We discover the need to go one-step closer from NLP-centered text classification towards integrating clinical psychology. This induces a call for in-depth analysis of users' perception behind their emotional intent in texts. With this background, we explore psychological theories for perception mining and causal analysis on social media in the past.

## 2.1. Related Work

A comprehensive literature on clinical psychology theories suggests possible detection of potential reasons behind mental disturbance in post that refer to insomnia, weight gain, or other indicators of worthlessness or excessive or inappropriate guilt. Underlying reasons may include *bias or abuse* Radell, Abo Hamza, Daghustani, Perveen and Moustafa (2021), loss of *jobs or career* Mandal, Ayyagari and Gallo (2011), physical/emotional illness leading to, or induced by the use of *medication* Smith (2015); Tran, Ho, Ho, Latkin, Phan, Ha, Vu, Ying and Zhang (2019), *relationship* dysfunction such as marital issues Beach and Jones (2002),[6] and *alienation* Edition et al. (2013). This list is not exhaustive, but it is a starting point for our study, giving rise to five categories of reasons for multi-class causal categorization.

In reference to our comprehensive survey on mental health analysis in social media posts, we witness the need of inspecting past studies for causal analysis in this domain Garg (2023). As per our problem definition, Son, Bayas and Schwartz (2018) performs partial studies with cause detection and causal explanation extraction from Facebook posts using Long Short Term Memory (LSTM) model highlighting availability of causality dataset over social media with an intensive iterative process of annotation. To handle this problem of lack of datasets, Tan, Zuo and Ng (2022) reveals causal event detection dataset and defines a novel NLP-centered approach of finding correlation among texts that belongs to news category. Their preliminary step of retaining samples with <= 3 sentence length limits their studies to short well-formed texts. In extension to their previous work on using LSTM to find discourse arguments, Son and Schwartz (2021) employs PDTB-style method that predicts discourse relation between adjacent discourse arguments of Tweets to capture the context across all other discourse arguments by using hidden vectors of argument pairs through *discourse argument-based LSTM*.

A data-driven approach uses Point-wise Mutual Information (PMI) to find correlations between two *verb phrases* acquired from data Chambers and Jurafsky (2008). Causal inference in place of correlation gives better and directional insights among different phrases Weber, Rudinger and Van Durme (2020). However, to the best of our knowledge, there is no other causal analysis of mental disturbance on Reddit data. The major challenges with Reddit dataset are length of post, in-formal user-generated texts containing slang and abbreviations. We come across long texts in real-time Reddit posts encircling life experiences. We employ end-to-end transformer-based models to propose solution for causal categorization in long texts.

---

[6] Freud (1957) develops a psycho-dynamic theory to find associations between depression and fear of losing loved ones.

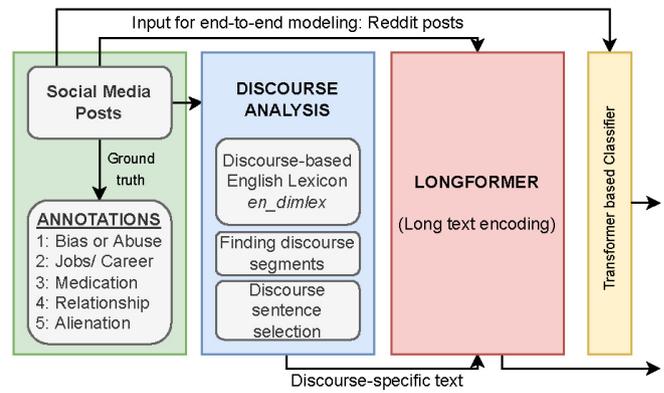

**Figure 2:** Block diagram of different modules for proposed approach to find possible solutions for multi-class causal categorization in long texts.

## 2.2. Problem Formulation

Consider a set of *n* number of candidate posts $P = \{P_1, P_2, P_3, ..., P_n\}$ in a given corpus. We employ cause-based categorization (causal categorization) of classifying any given $i^{th}$ post $p_i$ into one of the predetermined categories $C = \{C_0, C_1, C_2, C_3, C_4\}$ where $C_0$: 'bias or abuse', $C_1$: 'Jobs and careers', $C_2$: 'Medication', $C_3$: 'Relationship', and $C_4$: 'Alienation' as *causal category labeling*: $C_{p_i}$. For instance, consider a following post $P_1$:

> $P_1$=It has been 5 years now when that horrible incident of ragging shattered down all my confidence.

Here, the post $P_2$ belongs to category $C_0$: *Bias or Abuse* because the user is upset with a ragging incident which turns out to be an abusive experience of their life. Thus, $C_{p_1} = C_0$. The crawled data from subreddits: r/depression and r/SuicideWatch reflect mental disturbance as intent of a user. M-CAMS corpus contains the candidate posts with causal text segments behind their intent. Thus, all given instances contain text segments carrying causal explanation of user's mental disturbance.

## 3. Method and Materials

We deploy end-to-end transformer-based models as multi-class classifiers for causal categorization. We apply Longformer as a potential tokenizer to encode long texts to handle the problem of long texts as evident from Table 2. We find the effectiveness of using Longformer for text encoding in comparison of traditional end-to-end BERT-based models. We further apply rule-based discourse analysis to select sentences with discourse segments and perform ablation studies. In this section, we delineate the proposed approach as shown in Figure 2.

### 3.1. Data Collection

We use existing SDCNL data Haque, Reddi and Giallanza (2021), and crawl Reddit posts through the Python

| Causal category (↓) | Training | Testing | Total |
|---|---|---|---|
| Bias or Abuse | 297 | 53 | 350 |
| Jobs and career | 484 | 144 | 628 |
| Medication | 510 | 113 | 623 |
| Relationship | 1070 | 272 | 1342 |
| Alienation | 1119 | 289 | 1408 |
| Total | 3480 | 871 | 4351 |

**Table 3**
Number of training, testing and total samples in the given corpus of candidate posts.

Reddit API Wrapper (PRAW)[7] to increase the number of posts upto 5051 instances. The dataset is given as CAMS dataset which defines two tasks, namely, causal categorization and causal explanations. There are six categories, including *no-reason category*, in original CAMS dataset as explained before. However, recent developments in causal analysis suggests the existence of cause-and-effect relationship in a given text for identifying text segments indicating reason behind mental disturbance. Thus, we modify the task of causal categorization by classifying it into two tasks: cause detection and causal categorization. Thus, we remove the samples with no-reason from original CAMS dataset and define new causal categorization task with remaining candidate posts that belongs to one of the five causal categories. This gives rise to a new dataset for Multi-task causal analysis dataset (M-CAMS): cause detection, causal categorization and causal explanation. The statistics about size of dataset is given in Table 3. The M-CAMS dataset for causal categorization consists of 5051 posts for cause detection and remove the posts with 'No reason/ cause' to obtain 4352 posts for causal explanation and causal categorization. We use standard training - testing split mechanism for experimental results. For annotations, a senior clinical psychologist in association with a rehabilitation councillor train three post-graduate students for annotating cause detection, causal category and causal explanations in the dataset.

### 3.2. Inter-Annotator Agreement

We carry out two-fold mechanisms for the annotation task on : (i) individual-labeling, and (ii) group-wise labeling. Among three tasks of causal analysis: cause detection, causal explanation, and cause categorization, we combine cause detection and causal categorization by stuffing the predefined five causal categories with an additional category as {0: No cause/ reason}, for individual-labeling[8]. We keep causal explanations for group-wise labeling[9]. Our verify and assign final annotation to carry out *Fliess' Kappa inter-annotator agreement (IAA) study*. Despite the increased subjectivity of causal detection and causal categorization task, the annotators agree 61.28% among themselves which suggests

---

[7] https://praw.readthedocs.io/en/stable/
[8] Each student annotator labels the corpus for one of the predefined categories: {0: No reason/ cause, 1: Bias or Abuse, 2: Jobs and career, 3: Medication, 4: Relationship, 5: Alienation}
[9] Three student annotators work in a group to extract the text-spans with causal explanation

*substantial agreement* as shown in Table 4. Low levels of inter-annotator agreement is a well known problem in emotion based subjective studies, where lower agreement scores are reported Tsakalidis, Papadopoulos, Voskaki, Ioannidou, Boididou, Cristea, Liakata and Kompatsiaris (2018) suggesting highly complex nature of the task. To our surprise, the IAA carried out by our experts for causal explanation extraction shows higher agreement with explanations suggesting the significance of causal analysis. The overarching idea of IAA for causal explanation extraction is the group-wise labeling of causal explanation for five predefined categories followed by twofold IAA by our experts: (i) 2 categories (agree, disagree) and (ii) 4 categories (strongly agree, weakly agree, weakly disagree, strongly disagree). We observe the values of $\kappa$ for 2 categories and 4 categories as 74.63% and 67.23%, respectively. Furthermore, we remove the data points with no-reason and perform experiments on 4352 posts.

We further observe the disagreement of annotators for specific classes. We observe more than 34% of major disagreements in the ratio of 2:1 and about 4% of major disagreements among all three annotators. The former disagreements are resolved with majority voting consensus and the later ones are verified by our senior clinical psychologist. Our fine-grained analysis shows higher disagreements in following cases:

1. **Multiple Causes**: Overlapping judgment on two different causes present in a given data point. According to clinical psychology, interpretation of causes are subjective. Consider the following post $P_1$:

   > $P_1$: I cannot do anything without screwing it up, I just got suspended from school. My family, my friends, all lose their trust in me, I'm just not cut out for anything I don't think the world has a place for me.

   In a given post $P_1$, user is upset about the education (*Jobs and career*) and social relations (*Relationship*). One interpretation is about suspension from school as the root cause suggesting *jobs and career*. The other interpretation is that mental disturbance is due to friends and family members losing trust in the user.

2. **Ambiguity in human interpretations**: Human interpretations of the same post and the same inference may vary, even among experts. Consider the following post P2:

   > P2: I wish I could stay alone somewhere and cry my self to sleep. I wish i won't wake up.

   P2 contains some important words like *alone* and *cry* which convey the category as *Alienation*. However, two out of three annotators considered this to be the *No reason* category. As this is the subjective decision of every human annotator, we leave it at their discretion.

3. **Relationship** and **Jobs and Career**: People who matters the most are an integral part of users interpersonal relations. Impact of near and dear ones on users' professional life may significantly affect mental disturbance. Considering the consequence as mental disturbance, the direct cause is interpreted as *jobs and career*. However, the consequence of mental disturbance in Jobs and career, suggests the reason to be near and dear ones or *relationships*.

| Range | Interpretation |
|---|---|
| < 0: | Less than chance agreement |
| 0.01–0.20: | Slight agreement |
| 0.21– 0.40: | Fair agreement |
| 0.41–0.60: | Moderate agreement |
| 0.61–0.80: | Substantial agreement |
| 0.81–0.99: | Almost perfect agreement |

**Table 4**
Interpretation of resulting values of Fliess' Kappa agreement study McHugh (2012).

4. **Bias or Abuse** and **Relationships**: People usually lose self-esteem due to abusive relationships and other associated biases. This injustice affects more often if the person who is impacting is a closed one. Body shaming or emotional abuse is caused by friends, family or intimate partners. In this scenario, the reason behind mental disturbance could be the *bias or abuse*, or a *relationship*.

5. **Bias or Abuse** and **Jobs and Career**: Sometimes, the reason behind frustration and work life stress is due to the unethical promotions and credits given to another colleague, unhealthy work environment and insufficient funds. Such biases could be due to many reasons of *ego development*[10]. In this scenario, decisions about the actual cause behind mental disturbance could be subjectively different.

### 3.3. Rule-based Discourse Analysis

One possible solution to find cause-and-effect relation in a given text is discourse analysis. A preliminary step of finding discourse segments as potential indicators of causal explanations may contribute towards improvement in end-to-end classifiers. As suggested by Son et al. (2018), we follow their standard rules to propose a rule-based discourse segment detection starting with Part of Speech tagging (POS) tagging. We first deploy POS tagging mechanism for social media posts followed by identifying connectives. Recently, Das, Scheffler, Bourgonje and Stede (2018) introduce Dimlex Lexicon as English discourse markers which we use to find connectives in given text.[11] Each sentence in a given text is split into different segments over discourse connectives. We further find association among connecting segments through POS tags and ensure that connection is formed between two activities rather than nouns. We select the activity based sentences and finally concatenate them for input to classifier.

---

[10]Ego development is the infant's emerging consciousness of being a separate individual distinct from others, particularly the parents.
[11]https://github.com/discourse-lab/en_dimlex

Discourse connectives are linguistic means whose function in a text is to signal semantic and rhetorical discourse relations. As suggested by Son et al. (2018), we test and validate Rule-based Discourse Analysis (RDA) as potential module to pre-process long texts for causal categorization. Not every sentence contains discourse connectives to represent argumentative nature of the post and RDA on such posts results in empty string. To this end, we propose a *Biased Rule-based Discourse Analyiss (B-RDA* which employs RDA on posts having $length > 200$. We keep discourse analysis as an optional module, and perform ablation studies with and without RDA and B-RDA.

### 3.4. Transformer-based Models: Encoding Long Texts

The most widely accepted Tokenizer for a given text is AutoTokenizer in the end-to-end transformer. The original BERT model was primarily designed to handle sentence-length and paragraph-length inputs and has a maximum input length of 512 word piece tokens, or about $200 - 400$ word tokens Gao, Alawad, Young, Gounley, Schaefferkoetter, Yoon, Wu, Durbin, Doherty, Stroup et al. (2021). In a given corpus, the data points with number of words more than 200, 300 and 400 is 1418 (32.57%), 815 (18.72%) and 531 (11.96%), respectively. This substantial number of long-texts in real-time scenarios arise the problem of length inconsistency among data points. The solution for length inconsistency in Reddit posts are two-fold: (i) removing long texts from corpus, (ii) encoding long texts. In this work, we choose to handle long texts through XLNet, MentalBERT and Longformer. Longformer's attention mechanism is a drop-in replacement for standard self-attention and combines the local windowed attention with task motivated global attention.

### 3.5. The use of Longformer

**The use of Longformer**: We use end-to-end multi-class classifiers. Original text is cleaned by removing URL and irrelevant units to preserve 'UTF-8' encoding of text. We further use tokenize mechanism with longformer pretrained model[12] and fine-tune Longformer with M-CAMS dataset for causal categorization. We witness the use of longformer on different tasks such as question answering, text classification and many more. However, we test the significance of pre-trained masked language modeling for categorizing the cause in long text as compared to BERT-based models. We determine the effectiveness of RDA and B-RDA for both BERT-based models and longformer.

The Longformer[13] is a large language model leveraging on the GPT (Generative Pre-trained Transformer) architecture. Longformer is trained on a much larger dataset and fine-tuned to classify new, unseen texts. The major advantage of Longformers is its ability to understand and generate long texts through linear scaling of sequence length in comparison of the quadratic scaling by pretrained Transformers.

---

[12]allenai/longformer-base-4096
[13]https://huggingface.co/allenai/longformer-base-4096

The transformer architecture uses self-attention mechanisms to weigh the importance of different parts of the input text and uses this information to make predictions. On the contrary, the Longformer adhere to windowed local-content self attention through sparse attention pattern which avoids quadratic computing with full matrix multiplication, followed by a global attention that encodes inductive bias about the classification task.

We extend the maximum input sequence length to 4,096 tokens, which is 8X of the conventional transformer-based models by introducing the localized sliding window and global attention mechanisms to reduce the computational expenses of full self-attention mechanisms $O(n^2)$. Like BERT and RoBERTa, Longformer uses the extra positional embedding for each token position which is increased from 512 for RoBERTa to the maximum input length of 4,096 tokens. We use the pretrained Longformer with two-fold projections: (1) Sliding window attention $(Q_s, K_s, V_s)$, and (2) Global Attention $(Q_g, K_g, V_g)$ through Equation 1.

$$Attention(Q, K, V) = softmax\left(\frac{QK^T}{\sqrt{d_k}}\right)V \quad (1)$$

To mitigate the problem of expensive operation through matrix multiplication $QK^T$, the dilated window attention computes only fixed number of diagonals of $QK^T$. We used the default classification model pretrained using a simple binary cross entropy loss on the top of the first [CLS] token and global attention to [CLS]. The model was trained with Adam optimizer with batch sizes of 32 and linear warmup and decay with warmup steps = 0.1 of the total training steps. Experiments were performed on a single RTX8000 GPU with learning rate of 3e-5 and 15 epochs set after grid search mechanism Beltagy, Peters and Cohan (2020). While most documents in our dataset are short, 18.72% of the documents are larger than 400 word-pieces. As per recommendations of Beltagy et al. (2020), we place the global attention on [CLS] token. Thus, to resolve the problem of truncating the given input in desired length of sequence by pretrained transformers, we apply Longformer which is able to build contextual representations of entire context using multiple layers of attention.

## 4. Experimental Results and Evaluation

In this section, we provide information on experimental set-up, hyperparameter optimization, baselines, and evaluation metrics/ protocols. We perform in-depth analysis of experimental results with comparative analysis, cause-based analysis, computational analysis and ablation studies.

### 4.1. Experimental Setup

The experiments are performed over causal categorization task of M-CAMS datasetIt is evident from statistics in Table 3 that number of samples for each cause varies in the range of [29, 1119] and [53, 289] for training and testing datasets, respectively. In our previous experiments with machine learning and deep learning, we resolve the problem of imbalanced dataset by selectively accommodating crawled instances in SDCNL dataset Garg et al. (2022). In this work, we choose pre-trained models to diminish the effect of inconsistency in the number of samples for each causal category. With the provided GPU on Google Colab, we train our models with varying values of batch size due to small sample size of dataset. In near future, we plan to explore few shot sampling based mechanisms such as prompt-learning or data augmentation for added samples.

### 4.2. Hyperparameter Optimization

We make use of default configuration and hyperparameters of pre-trained Huggingface's models.[14] For consistency, we use same experimental settings for all models with 10 fold cross-validation and results are reported as average of all folds. We use a grid search optimization technique to optimize parameters. To tune the number of layers ($n$), we empirically experiment with the values: $n \in \{1, 2, 3\}$. For drop out ($\delta$), we test it with: $\delta \in \{0, 0.2, 0.4, 0.6, 0.8\}$, hidden dimension ($H$) with H $\in \{64, 128, 256\}$, learning rate (lr): lr $\in \{1e-5, 3e-5, 5e-5\}$. For optimization (O): O $\in \{$ 'Adam', 'Adamax', 'AdamW'$\}$ with a batch-size of $\{8, 16, 32, 64\}$ were used. We vary the number of epochs as $\{5, 10, 15\}$.

We observe the best performance by varying batch-size and learning rate for Longformer as shown in Table 5. On further investigating the loss incurred during testing phase and accuracy so obtained with varying values of hyperparameters as shown in Table 6, we find the highest accuracy with lr = 1e-5 and batch size= 32. In addition to this, we show the time taken by each iteration in Table 7 and observe that more training time is required for lower values of batch-size. After in-depth analysis of trade-off between efficiency of the model and time complexity, we find best performance with lr = 1e-5 and batch size= 32. Thus, we set hyperparameter for our experiments as $n = 3$, $\delta = 0.8$, $H = 256$, $O =$ Adam, lr = 1e-5, batch size= 32 and epochs=5.

### 4.3. Baselines

We compare some well-known baseline models in literature for the task of causal categorization:

1. **BERT**: BERT is a bidirectional transformer for language understanding Devlin, Chang, Lee and Toutanova (2018) which is pre-trained using a combination of masked language modeling objective and next sentence prediction on a large corpus comprising the Toronto Book Corpus and Wikipedia.[15] The use of BERT for multi-class classification is outperformed by domain specific transformers.

2. **DistilBERT**: DistilBERT Sanh, Debut, Chaumond and Wolf (2019) was pre-trained on raw texts only, with no human labelling and can use publicly available data for fine-tuning to automate generating inputs and labels using BERT base model.[16]

---
[14] https://huggingface.co/models
[15] https://huggingface.co/docs/transformers/model_doc/bert
[16] https://huggingface.co/distilbert-base-uncased

| Causal Category (↓) | Batch Size (→) | 8 | | | 16 | | | 32 | | | 64 | | |
|---|---|---|---|---|---|---|---|---|---|---|---|---|---|
| | lr (↓) | P | R | F | P | R | F | P | R | F | P | R | F |
| Bias/ Abuse | | 0.32 | 0.29 | 0.30 | 0.23 | 0.38 | 0.28 | 0.11 | 0.50 | 0.18 | 0.13 | 0.58 | 0.22 |
| Jobs/ Career | | 0.47 | 0.70 | 0.56 | 0.65 | 0.68 | 0.56 | 0.56 | 0.66 | 0.61 | 0.69 | 0.59 | 0.64 |
| Medication | $1e-5$ | 0.51 | 0.47 | 0.49 | 0.61 | 0.48 | 0.54 | 0.50 | 0.52 | 0.51 | 0.51 | 0.49 | 0.50 |
| Relationship | | 0.64 | 0.58 | 0.61 | 0.60 | 0.66 | **0.63** | 0.71 | 0.65 | **0.68** | 0.63 | 0.65 | 0.64 |
| Alienation | | 0.62 | 0.61 | **0.62** | 0.64 | 0.60 | 0.62 | 0.70 | 0.60 | 0.65 | 0.67 | 0.63 | **0.65** |
| **Testing Accuracy** | | | 0.57 | | | 0.60 | | | **0.62** | | | 0.61 | |
| Bias/ Abuse | | | | | 0.43 | 0.32 | 0.37 | 0.08 | 0.50 | 0.13 | 0.00 | 0.00 | 0.00 |
| Jobs/ Career | | | | | 0.61 | 0.56 | 0.59 | 0.58 | 0.66 | 0.62 | 0.42 | 0.71 | 0.53 |
| Medication | $3e-5$ | | | | 0.50 | 0.49 | 0.49 | 0.48 | 0.50 | 0.49 | 0.59 | 0.52 | 0.55 |
| Relationship | | | | | 0.60 | 0.64 | 0.62 | 0.56 | 0.68 | 0.61 | 0.70 | 0.64 | 0.67 |
| Alienation | | | | | 0.57 | 0.61 | 0.59 | 0.74 | 0.53 | 0.62 | 0.73 | 0.59 | 0.65 |
| **Testing Accuracy** | | | 0.33 | | | 0.57 | | | 0.58 | | | 0.61 | |

**Table 5**
Performance evaluation of longformer over different causal categories for varying values of learning rate (lr) and batch size.

| Batch Size (→) | 8 | | 16 | | 32 | | 64 | |
|---|---|---|---|---|---|---|---|---|
| lr (↓) | Loss | Accuracy | Loss | Accuracy | Loss | Accuracy | Loss | Accuracy |
| $1e-5$ | 1.1524 | 0.5695 | 1.1360 | 0.5993 | **1.0649** | 0.6177 | 1.0765 | 0.6073 |
| $3e-5$ | 1.4745 | 0.3318 | 1.1798 | 0.5706 | 1.1657 | 0.5844 | 1.0909 | 0.6073 |
| $5e-5$ | 1.4733 | 0.3318 | 1.4768 | 0.3318 | 1.2105 | 0.5729 | 1.4722 | 0.3318 |

**Table 6**
Loss and Accuracy of longformer for varying values of learning rate (lr) and batch size.

3. **RoBERTa**: RoBERTa is a Robustly Optimized BERT Pretraining Approach Liu, Ott, Goyal, Du, Joshi, Chen, Levy, Lewis, Zettlemoyer and Stoyanov (2019) which is build on BERT and modifies key hyper-parameters, removing the next-sentence pretraining objective and training at much larger mini-batches and learning rates.[17]

4. **XLNet**: The XLNet Yang, Dai, Yang, Carbonell, Salakhutdinov and Le (2019) is a generalized autoregressive pretrained model for language understanding employs Transformer-XL as the backbone model, exhibiting excellent performance for language tasks involving long context.[18]

5. **MentalBERT**: MentalBERT Ji, Zhang, Ansari, Fu, Tiwari and Cambria (2021) is trained with mental health-related posts collected from Reddit.[19]

[17] https://huggingface.co/docs/transformers/model_doc/roberta
[18] https://huggingface.co/xlnet-base-cased
[19] https://huggingface.co/mental/mental-bert-base-uncased

### 4.4. Evaluation metrics and protocols

We evaluate the performance of different transformer based models with standard performance evaluation: precision, recall and f-measure. We find suitable of models with minimum loss and highest accuracy to trace the value of loss during training and testing. Analyzing trade-off between loss and accuracy analysis helps in determining the optimum model. We investigate training accuracy, inference accuracy, macro accuracy, and weighted accuracy of all models. *Macro accuracy* computes F1-measure for each class, and returns the average without considering the proportion of each class in a dataset. *Weighted Accuracy* compute F1-measure for each class, and returns the average considering the proportion of each class in a dataset. We use statistical significance test of Mann–Whitney U test ($p < 0.5$) to test statistical significance of longformers.

### 4.5. Comparative Analysis

We perform comparative analysis of transformer-based models as shown in Figure 3. As evident from F-measure, RoBERTa shows minimal performance with *bias/ abuse category* but shows improved performance with precision

| Batch Size (→) | 8 | | | 16 | | | 32 | | | 64 | | |
|---|---|---|---|---|---|---|---|---|---|---|---|---|
| LR (↓) | **Train.** | **Val.** | **Inf.** | **Train.** | **Val.** | **Inf.** | **Train.** | **Val.** | **Inf.** | **Train.** | **Val.** | **Inf.** |
| $1e-5$ | 182 | 5 | 5 | 145 | 6 | 5 | 129 | 5 | 5 | 119 | 5 | 5 |
| $3e-5$ | 190 | 6 | 6 | 143 | 5 | 5 | 130 | 7 | 5 | 120 | 6 | 5 |
| $5e-5$ | 183 | 5 | 5 | 144 | 5 | 5 | 130 | 6 | 5 | 119 | 5 | 5 |

**Table 7**
Time taken (in approximate seconds) by longformer for varying values of learning rate (lr) and batch size.

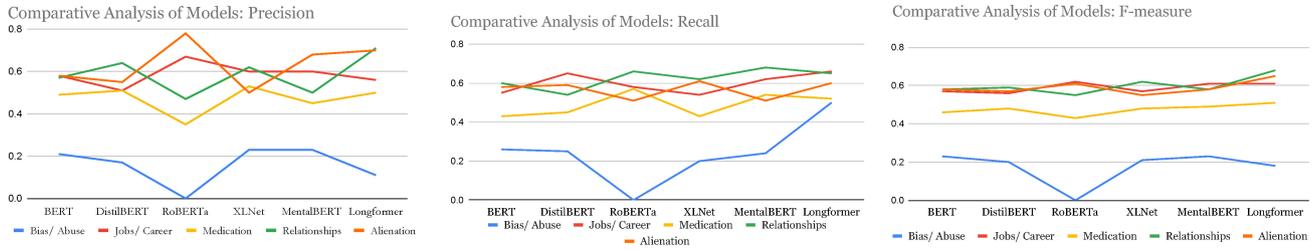

**Figure 3:** Comparative analysis of transformer-based models in F-measure, Precision and Recall.

| Methods (↓) | Precision | Recall | F-measure | Accuracy |
|---|---|---|---|---|
| BERT | 0.486 | 0.484 | 0.484 | 0.544 |
| DistilBERT | 0.476 | 0.496 | 0.480 | 0.543 |
| RoBERTa | 0.454 | 0.464 | 0.442 | 0.560 |
| XLNet | 0.496 | 0.480 | 0.486 | 0.542 |
| MentalBERT | 0.492 | 0.518 | 0.498 | 0.554 |
| Longformer | **0.516*** | **0.586*** | **0.526*** | **0.618*** |

**Table 8**
Comparative analysis of all multi-class categorization models for causal categorization task of M-CAMS dataset. F1, Precision, and Recall scores are averaged over 10 folds. *indicates that Longformer achieved a significant ($p < 0.05$) performance improvement over best baseline under Mann–Whitney U test for *medication*. This supports RoBERTa as outperforming model for medical social media analysis as evident from past studies Casola and Lavelli (2020). In contrast, we observe highest recall with Longformer for *bias/abuse*. Considering performance with domain-specific social media data, MentalBERT gives third best performance. To this end, in near future, it will be worth investigating longformer tuned over domain-specific context-aware models.

*Recall* suggests the number of correct predictions for each class made out of all samples of corresponding class in a dataset. In contrast to recall, *precision* reports the number of correct predictions made for labels predicted for each class in a dataset. The stakeholders such as clinical psychologists wish to find the correct cause for a given sample highlighting the importance of underlying recall more than precision in this task of multi-class causal categorization. Determining close proximity of values in *recall* measure therefore induce better insights as compare to *precision*. We clearly observe close proximity of recall values in Longformer that outperforms all other transformer-based models.

In practice, for multi-class classification model accuracy is mostly favored over f-measure. We obtain precision, recall, f-measure and accuracy for all the models as shown in Table 8 and high values shows the efficiency of Longformer against other models. However, it makes more sense to analyse F1-measure due to unbalanced sample size among different classes. On examining accuracy, Longformer shows significant improvement in comparing transformer-based methods through Mann-Whitney U test for $p < 0.5$. We further look into performance evaluation through *training accuracy,* *inference accuracy, macro accuracy, and weighted accuracy* as shown in Figure 5.

The least variation in training and inference accuracy is observed for DistilBERT and Longformer as shown in Figure 5. The Longformer further outperforms DistilBERT and all other baselines with fair margin as evident from graph plot for testing accuracy. For our task, weighted accuracy is more meaningful than macro accuracy as we observe difference in number of samples for each class in testing data. Henceforth, we observe weighted accuracy where Longformer outperforms the baselines followed by RoBERTa. The increase in values of weighted accuracy as compare to macro accuracy is because of its inclination towards more number of samples for causal categories with better f-score such as *relationship* and *alienation*.

### 4.6. Cause-based Analysis

Based on five different categories, we evaluate performance for cause-based analysis as shown in Figure 4. The value of precision varies maximum for *Alienation* and minimum for *bias/ abuse*. On the other hand, the values of recall varies maximum for *bias/ abuse* and minimum for *alienation* which shows correlation with minimum and maximum data size for *bias/ abuse* and *alienation*, respectively. Longformer outperforms 4 out of 5 categories and shows comparable performance with *bias/ abuse*. The resulting scores of *bias/ abuse* and *medication* are lowest as compared to highest scores for *relationship* and *alienation*. These results validates the opinion stated by experts (social NLP researchers and a senior clinical psychologists).

Social NLP researchers suggests that NLP in medical texts is still at premature stage to map keywords in texts reflecting medication based mental disturbance. Clinical psychologists reveals the common pattern in writing of users about relationships and alienation as society is still immature enough to handle relations and isolation related issues. Therefore, the words that users use to express such problems incur information from their active vocabulary such as *father, mother, sister, girlfriend, husband* for *relationship* category and *alone, getting bored, emptiness* for *alienation category*.

Moreover, the major cause behind misinterpretation of bias/abuse category could be due to low agreement score with dilemma in category labeling: *bias/ abuse, relationship and alienation*. The annotation perplexity guidelines resolve this dilemma during human judgments. Comparative low performance for bias/abuse category as compare to all other categories validates the need of context-aware transformer based models for categorizing cause in the text in near future.

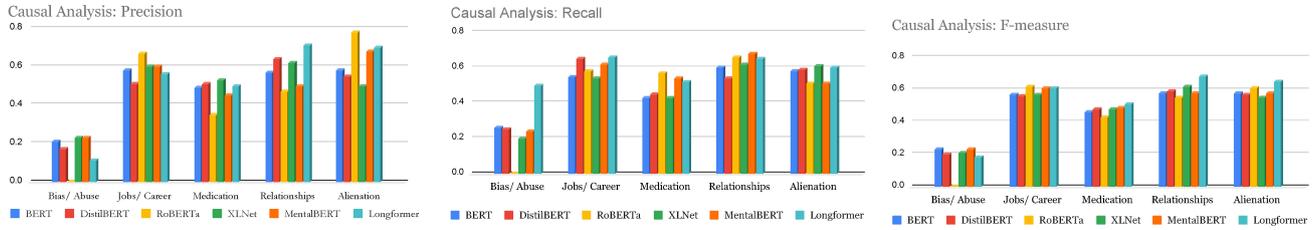

**Figure 4:** Performance evaluation of different methods for five causal categories of mental disturbance in terms of F-measure, Precision and Recall.

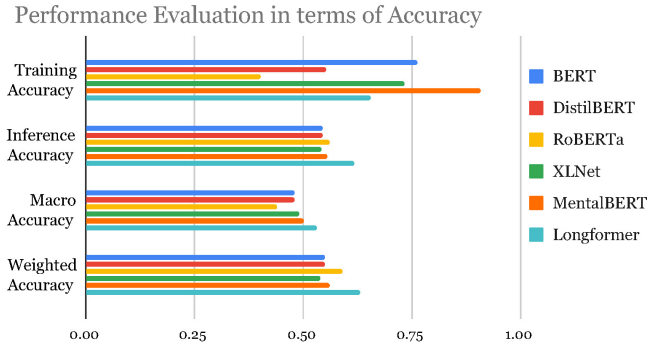

**Figure 5:** Comparing accuracy of Longformer v/s the baselines.

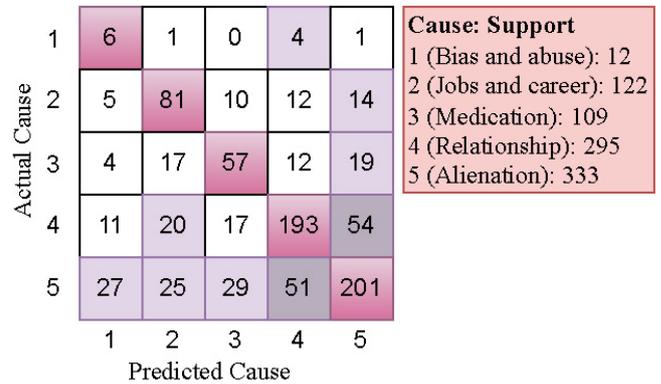

**Figure 6:** The confusion matrix for Analysis of causal categories where the diagonal axis represent correctly detected right values (True positives).

### 4.7. Error Analysis

The accuracy of the five-class classification task is found to be 60% with the Longformer. We undertake a comprehensive error analysis to explore the intricacies of our results as shown in Figure 6. We observe highest prediction numbers for 'Relationships' and 'Alienation' due to specific keywords observed for corresponding categories such as 'family, friends, broke up, party, fights' for *relationship*, and 'nothing to do, boring life, no motivation' for *alienation*. We observe similar nature for *jobs and career* which shows comparable accuracy with a comparatively lower number of supporting data points. Surprisingly, the highest number of mis-predictions are due to 'Relationships' and 'Alienation' followed by 'Jobs and career'.

As we observe the color code, the white boxes represent a low number of mis-predictions and dark grey boxes represent a high number of mis-predictions. The diagonal row represents more than 50% of true positives for every category. However, there are some pairs of classes which have more confusion than the others. Consider the following types of error:

1. **Overlapping of 'Alienation' and 'Relationship'**: The *overlapping problem* of classes is observed with ambiguous results for *Relationship* and *Alienation*, probably due to higher number of data points in the original dataset. As we observe from high disagreement between *Relationship* and *Alienation* in Section 3.2, the demarcation of fixed boundaries is not possible in real-time scenario due to subjectivity of the task. We recommend the approximation of a newly built model over handcrafted/ automated features accordingly. Consider the following post P1:

    > P1: My friends are ignoring me and I am left all alone, doing nothing. I have lost all my friends and living a boring life.

    The user is disturbed due to broken friendship impacting the social lifestyle of a person with *alienated* lifestyle. However, the underlying cause is broken friendship.

2. **Prediction of 'Bias or Abuse' as 'Relationship'**: We obtain low performance for *'Bias or abuse'* due to annotators' perceived overlap with 'Relationship'), for instance, the data points referring about an *abusive relationship*. Future work is needed to mitigate such uncertainty. For example, delineation of discourses within the text would support a more definitive interpretation and reliable annotation. Consider the following post P2:

    > P2: Phew! I am so fat. Now, it is affecting my relationship with my wife. She does not find me attractive anymore and is cheating on me. I give up.

| Methods (↓) | Train. | Inf. | Loss |
|---|---|---|---|
| BERT | 725 | 16 | 1.8299 |
| DistilBERT | 9418 | 5 | 1.5265 |
| RoBERTa | 912 | 18 | 1.2802 |
| XLNet | 1120 | 20 | 1.5237 |
| MentalBERT | 798 | 16 | 2.2306 |
| Longformer | **129** | **5** | **1.0649** |

Table 9
Comparing time taken and loss incurred by Longformer v/s baselines.

| Methods (↓) | A | M-F1 | W-F1 | Loss |
|---|---|---|---|---|
| Transformer | 0.56 | 0.44 | 0.59 | 1.28 |
| RDA + Transformer | 0.51 | 0.30 | 0.58 | 1.30 |
| B-RDA + Transformer | 0.48 | 0.24 | 0.59 | 1.26 |
| Longformer | **0.62** | **0.53** | **0.63** | **1.06** |
| RDA + Longformer | 0.58 | 0.50 | 0.58 | 1.18 |
| B-RDA + Longformer | 0.54 | 0.42 | 0.57 | 1.24 |

Table 10
Ablation study: A: Accuracy, M-F1: Macro Accuracy, W-F1: Weighted Accuracy; RDA: Rule based Discourse Analysis, B-RDA: Biased - Rule based Discourse Analysis, LF: Longformer

The text-spans refers to an explanation about the body shaming affecting their relationship and hence, the mental health. However, the underlying cause is physical abuse.

3. **Prediction of 'Relationship' as 'Jobs and career'**: The circumstances where *'relationships'* such as parent-child and intimate-partners triggers the problems in education and jobs, respectively. Such situational effect should be interpreted by infusing perplexity guidelines and knowledge through semantics in AI models effectively. Consider the following post P3:

   > P3: My mother does not want me to stay with her. She says that I am burden on her and I must do job and earn my own living. I miss her so much.

   In P3, although the words such as *job* and *earn* specifies *jobs and career*, we find the reason behind mental disturbance is distance from a strong relationship (mother-child).

4. **Overlapping of 'Jobs and career'/ 'Medication' and 'Alienation'**: The number of data points of 'Alienation', being the highest, skewed the prediction of other classes towards itself. There are some instances in *'Jobs and career'* and *'medication'* which are associated with boredom due to school and lazy due to physical misfit or inactivity, respectively. Such data points thus, manipulate the models towards *'Alienation'*. Consider the following post P4:

   > P4: I have lost everything in my life. Five years ago, after a car accident, I got bed-ridden and extremely lazy to do anything. I now feel like doing nothing, boring life and there no motivation.

   The underlying root cause of mental disturbance is the *accident* suggesting the causal category of *'Medication'*. However, words such as *'lazy', 'boring', 'no motivation'*, suggests *'Alienation'*.

To summarize this, we observe that the AI models are still not intelligent enough to find semantics of a given text to identify the root cause from the observed probabilities for each causal category.

### 4.8. Computational Analysis

Pre-trained model occupies large space and requires high computational time. In addition to this, cost analysis of methods in terms of loss values determines the effectiveness of a model. Loss is the penalty for a bad prediction. To this end, we examine computational time and loss for testing values under given experimental setup as shown in Table 9. The training time is minimal for Longformer and inference time is comparative to second fastest inferring model of distilBERT. Loss, being a non-beneficial attribute, is better with low values. The lowest value of loss is observed for Longformer. After in-depth analysis of computation time and cost, we determine the efficiency of Longformer.

### 4.9. Ablation study

Ablation studies damage and/or remove certain components in a controlled setting to investigate all possible outcomes of system failure. We employ discourse components and perform studies with and without discourse components: (i) rule-based discourse analysis (RDA), and (ii) biased rule-based discourse analysis (B-RDA) with our proposed model and determine the efficiency of modified version. A clear observation from Table 10 suggests the Longformer as best model and degradation in performance with RDA on long text narrated by social media users about their mental disturbance. This could be due to the fact that there might not be many instances of argumentative segments hindering the performance by reducing amount of information present in long texts. We therefore need context-aware models to accommodate the cause-and-effect relationship for pre-trained models.

## 5. Discussion

In the past, we carry out experiments with machine learning and deep learning models which are inefficient to solve this task as shown in our previous work Garg et al. (2022). To further enhance the observations and test the effectiveness of pre-trained models for multi-class causal categorization task, we investigate tranformer-based models empirically. We further employ rule-based discourse analysis as suggested by Son et al. (2018) and test it on causal categorization task of M-CAMS dataset.

### 5.1. Ethical considerations

To adhere to privacy constraints, we do not disclose any personal information such as demographics, location, and personal details of social media user while making this data available. We formulate guidelines for annotations which is carried out under observation of a senior clinical psychologist and rehabilitation councillor. This research is purely observational and we do not claim any solution for clinical diagnosis at this stage. The instances in CAMS data might subject to biased demographics such as race, location and gender of a user. Therefore, we do not claim diversity in CAMS dataset. In addition to this, we adhere to race, gender and ethnicity neutral conditions while writing this paper to follow an unbiased approach.

### 5.2. Open Challenges

We investigate a research direction of RDA and cause and effect relationship in long texts. After in-depth analysis of discourses over CAMS dataset, we find new research directions to investigate for solving multi-task causal categorization in long texts. We therefore pose new research directions as follows:

1. **Explainable AI**: In the past, Saxena, Garg and Ansari (2022) find explanations in the form of text segments in original post to find reasons behind causal categorization with deep learning models through LIME and Integrated Gradient methods. However, there exist need to find explanations for transformer-based classifiers to test the responsibility of AI models.

2. **Discourse Embedding**: We suggest investigating discourse embedding and pragmatics for syntactic analysis and semantic enhancements in a given text. We plan to explore knowledge graphs and cause-based summarization from long texts in near future.

3. **Data Augmentation**: We plan to explore cause-specific data augmentation to multiply the number of instances and investigate the effect of data augmentation on classifier and explainability.

## 6. Conclusion

We finally conclude our work by introducing transformer-based models and useful insights on possible improvements with discourse analysis. Introduction of M-CAMS dataset opens up new research directions toward multi-task causal analysis of mental disturbance in long texts. We find longformer and a best possible solution for causal categorization in user-generated long texts. Our proposed approach accommodates more than 98% of data and achieves acceptable accuracy of 62% on given data with 61.28% mutually agreed annotations through Fliess' Kappa inter-observe agreement study. We show the effectiveness of our model with cause-based analysis and ablation studies. Our work demonstrates the inefficiency of rule-based discourse analysis for identifying cause-and-effect relationship in user-generated complex and in-formal social media long-text through English Discourse markers.

## 7. Acknowledgement

We acknowledge a great consultation by Dr. Veena Krishnan, a senior clinical psychologist, and Dr. Ruchi Joshi, a rehabilitation counselor. We also acknowledge three student annotators: Simran jeet Kaur, Astha Jain and Ritika Bhardwaj.